\documentclass[letterpaper, 10 pt, conference]{ieeeconf}  

\IEEEoverridecommandlockouts                              

\usepackage{amsmath} 
\usepackage{amssymb}  
\usepackage{times}
\usepackage{scrextend}
\usepackage{mathtools}
\usepackage{algorithm}
\usepackage{algpseudocode}
\usepackage{lipsum}
\usepackage{cases}
\usepackage{multirow}

\usepackage{soul}
\usepackage{xcolor,colortbl}
\definecolor{LightCyan}{rgb}{0.88,1,1}

\usepackage{graphicx}
\usepackage{flushend}

\usepackage{color}
\usepackage{hhline}

\newcommand{\bq}{\mathbf{q}}
\newcommand{\btau}{\boldsymbol{\tau}}
\newcommand{\bomega}{\boldsymbol{\omega}}

\newcommand{\bnu}{\boldsymbol{\nu}}
\newcommand{\balpha}{\boldsymbol{\alpha}}
\newcommand{\bgamma}{\boldsymbol{\gamma}}
\newcommand{\boeta}{\boldsymbol{\eta}}
\newcommand{\bbeta}{\boldsymbol{\beta}}

\title{\LARGE \bf
 On the Performance of Jerk-Constrained Time-Optimal \\ Trajectory Planning for Industrial Manipulators }

\author{Jee-eun Lee$^{1}$, Andrew Bylard$^{2,*}$, Robert Sun$^{2}$, and Luis Sentis$^{1,*}$
\thanks{This work was supported by Dexterity, Inc.}
\thanks{* Corresponding authors contributed equally.}
\thanks{$^{1}$ J. Lee and L. Sentis are with The University of Texas at Austin, Austin, TX, USA, 
{\tt\small \{jelee, lsentis\}@utexas.edu}}%
\thanks{$^{2}$ A. Bylard, and R. Sun are with Dexterity, Inc., Redwood City, CA, USA, 
{\tt\small \{andrew.bylard, robert\}@dexterity.ai }}%
}

\begin{document}

\maketitle
\thispagestyle{empty}
\pagestyle{empty}

\begin{abstract}

Jerk-constrained trajectories offer a wide range of advantages that collectively improve the performance of robotic systems, including increased energy efficiency, durability, and safety. In this paper, we present a novel approach to jerk-constrained time-optimal trajectory planning (TOTP), which follows a specified path while satisfying up to third-order constraints to ensure safety and smooth motion. One significant challenge in jerk-constrained TOTP is a non-convex formulation arising from the inclusion of third-order constraints. Approximating inequality constraints can be particularly challenging because the resulting solutions may violate the actual constraints. We address this problem by leveraging convexity within the proposed formulation to form conservative inequality constraints. We then obtain the desired trajectories by solving an $\boldsymbol n$-dimensional Sequential Linear Program (SLP) iteratively until convergence. Lastly, we evaluate in a real robot the performance of trajectories generated with and without jerk limits in terms of peak power, torque efficiency, and tracking capability.  
\end{abstract}


\section{INTRODUCTION}

\begin{figure}[t!]
    \centering    \includegraphics[width=0.95\linewidth]{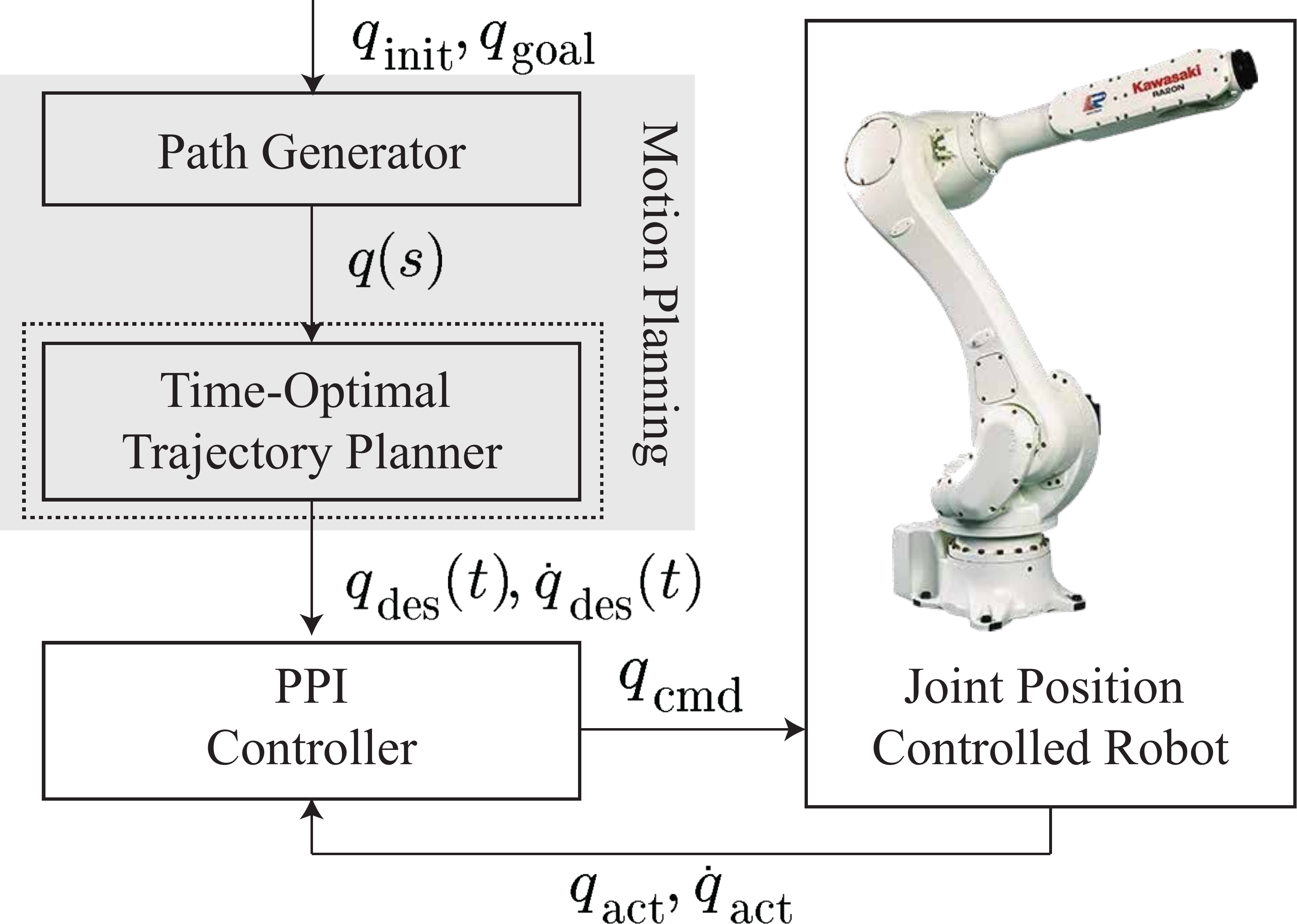}
    \vspace{-5pt}
    \caption{Two-Stage Motion Planning and Control Architecture: The path generator creates a collision-free path, which is fed into the time-optimal trajectory planner to compute the desired trajectory satisfying robot hardware limits. Subsequently, the predictive proportional-integral (PPI) controller generates high-frequency joint commands with feedback.}
    \label{fig:architecture}
    \vspace{-15pt}
\end{figure}

In automation, time-optimal motion planning plays a crucial role in maximizing productivity while ensuring efficiency and safety. However, solving the complete trajectory-planning problem is often difficult to perform in real time~\cite{chettibi2004minimum, diehl2006fast, schulman2013finding, zhao2018efficient, zhang2022time}. To obtain satisfactory results using reasonable computational resources, trajectory planning is often solved in two stages~\cite{shin1985minimum, shiller1989robot, shiller1991computing}. As shown in Fig. \ref{fig:architecture}, during the initial stage, the path generator creates a geometric path that incorporates task specifications and obstacle avoidance. In the subsequent stage, a time-optimal trajectory is calculated along the specified geometric path, enforcing constraints such as permissible joint velocity and torque. 

This paper focuses on solving the second-stage motion planning sub-problem, also known as time-optimal trajectory planning (TOTP) or path tracking (TOPT)~\cite{bobrow1985time, pfeiffer1987concept}. Although TOTP under first- and second-order constraints (such as velocity, acceleration, and torque limits) has been thoroughly explored, third-order constraints such as jerk or torque rate bounds are often overlooked. This is primarily because their inclusion lacks clear guidance (for instance, these bounds may not be specified in motor usage requirements), and they introduce non-convex terms into the formulation, making problem-solving much more challenging. However, without considering jerk limits, solution trajectories may include large acceleration jumps, resulting in various issues when applied to real robots~\cite{eager2016beyond}. 

In this paper, we propose a new approach for TOTP with practical third-order constraints. By incorporating jerk limits into trajectory planning and control strategies, robotic systems can achieve higher reliability and effectiveness. Some well-known advantages include: 1) Enhanced path tracking: a smoother trajectory allows the robot's movements to more closely match the desired trajectory, minimizing jerk-induced vibrations. 2) Reduced wear and tear: Sudden changes in acceleration can subject mechanical components to higher stress levels. By constraining jerk, the robot's motor gears undergo gentler transitions, extending component lifespans. 3) Energy efficiency: Jerk-limited motions prevent unnecessary spikes in power consumption, leading to decreased energy usage and reduced operational costs. 

Finally, we perform real robot experiments to confirm the benefits of including jerk limits. We analyze factors such as peak power and torque, which affect the performance, durability, and maintainability of the robot. The proposed motion planning and control framework is depicted in Fig.~\ref{fig:architecture}.

\subsection{Related Works}
Traditionally, there are two widely adopted approaches for addressing time-optimal trajectory planning (TOTP): numerical integration (NI), and convex optimization (CO). NI-based approaches~\cite{bobrow1985time, pfeiffer1987concept, pham2014general} solve the problem based on the insight that one can obtain a time-optimal trajectory through bang-bang control. However, this method requires identifying switch points where acceleration changes, and it is prone to failure in certain cases. Conversely, optimization provides a general and robust means to formulate and solve the problem~\cite{verscheure2009time, hauser2014fast}. However, it increases the problem's complexity by at least an order of magnitude, resulting in longer computation times. Other works solve the problem using dynamic programming~\cite{shin1986dynamic, singh1987optimality} or by leveraging specific conditions which may be too constraining, such as imposing constant first- or second-order limits in the path parameter space~\cite{haschke2008line, berscheid2021jerk}, which may not map to desired (e.g., constant) limits in the joint space.

Recently, there have been efforts to simplify TOTP algorithms through methods based in reachability analysis (RA)~\cite{pham2018new, consolini2019optimal}. These methods, instead of solving a high-dimensional optimization problem (with $n$ optimization variables, one for each timestep), can achieve the optimal solution by solving a series of 2$n$ 2-D sub-problems. However, it's worth noting that these reachability-based algorithms are restricted to second-order constraints, since they cannot handle situations where second-order constraints violate the third-order constraints, known as third-order singularity~\cite{pham2017structure}.

Many works have attempted to incorporate third-order constraints into time-optimal trajectory planning (TOTP). Many of them rely on the NI-based approaches~\cite{pham2017structure,ma2021new,mattmuller2009calculating, wang2021third}. In~\cite{ma2021new}, however, this formulation is limited to bounding the third-order derivative of the path parameter ($\dddot{s}$) rather than directly limiting the robot jerk or torque rate, and the third-order singularity issue is not addressed. Instead, one can carefully bridge the maximum and minimum jerk profile to avoid failure from singularity~\cite{pham2017structure}. However, this approach requires handling numerous exceptional cases, making the algorithm overly complex when adding first and second-order constraints, which are often more critical~\cite{wang2021third}.


Our approach provides a more general formulation within a convex optimization framework. A key challenge is enforcing nonlinear third-order inequality constraints, as approximating them without careful consideration can lead to solutions that violate real constraints. An alternative is to minimize both jerk and time~\cite{lu2017solving}, but this requires careful tuning of the relative weight for jerk and time in the cost function. Since the solutions of the optimization problem are often very sensitive to the weight, the solution can frequently sacrifice motion time or smoothness excessively depending on the input path given, even under the same weight. In this paper, we formulate the problem as an SLP using conservative linearization of the nonlinear constraints at each iteration to ensure solution feasibility. The work in \cite{zhang2013practical} also used conservative approximation for third-order constraints, based on the fact that torque obtained without third-order constraints is always bigger than the one with third-order constraints. Thus, their approach is limited to a one-time approximation, which can lead to suboptimal solutions, with stricter limits often widening the optimality gap.


\subsection{Contribution}
\begin{itemize}
    \item We formulate a unique TOTP with third-order constraints as a Sequential Linear Program (SLP), exceeding the performance of existing methods.
    \item We introduce a method for approximating general third-order constraints, enabling iterative optimization to converge to the optimal solution.
    \item We demonstrate that imposing jerk limits significantly reduces peak power, enhances energy efficiency, and improves tracking performance.
\end{itemize}

\subsection{Organization}
The paper is organized as follows. Section~\ref{sec:preliminaries} revisits the concepts of TOTP and the corresponding parameterization methods. Section~\ref{sec:jerktotp} formulates the jerk constraints for TOTP and outlines the overall optimization process with the approximated jerk constraints. In Section~\ref{sec:experiment}, we discuss the numerical and experimental results of the proposed jerk-bounded time-optimal trajectory planning. Finally, in Section~\ref{sec:conclusion} we discuss conclusions and future work.

\section{PRELIMINARIES}
\label{sec:preliminaries}
\subsection{Problem Formulation}
We formulate time-optimal trajectory planning (TOTP) as follows: Given a path $\bq(s)$ as a function of a scalar path coordinate $s\in[0,1]$, find a monotonically increasing time scaling $s(t) :[0,T] \rightarrow [0,1]$ that 
\begin{itemize}
    \item satisfies an initial state $(s_0,\dot{s}_0)=(0,0)$ and a final state $(s_\textrm{end},\dot{s}_\textrm{end})=(1,0)$, 
    \item minimizes the total travel time $T$ along the path,
    \item enforces kinematics and dynamics constraints imposed by robot hardware limitations.
\end{itemize}
\subsection{Time-Optimal Trajectory Planning (TOTP) }
For the purpose of solving the TOTP problem to track the prescribed path $\bq(s)$ for a robot manipulator, we express the first- and second-order time derivatives of $\bq(s)$ as
\begin{eqnarray}
\dot{\bq}(s) = \bq'(s)\dot{s}, \quad \ddot{\bq}(s)=\bq''(s)\dot{s}^2+\bq'(s)\ddot{s}.
\end{eqnarray}
We also transform the dynamics of the manipulator into
\begin{align}
\btau &~= M(\bq)\ddot{\bq} + C(\bq, \dot{\bq})\dot{\bq} + g(\bq)  \nonumber \\
 &~= M(\bq)(\bq''\dot{s}^2+\bq'\ddot{s}) + C(\bq, \bq')\bq' \dot{s}^2 + g(\bq)  \nonumber \\
&:= \mathbf{m}(s)\ddot{s} + \mathbf{c}(s)\dot{s}^2 + \mathbf{g}(s) \label{eqn:parameterized_dyn}.
\end{align}
Note that from the linear nature of the Coriolis matrix, we have $C(\bq, \bq'\dot{s})\bq' \dot{s} = C(\bq, \bq')\bq' \dot{s}^2$ above. 

Furthermore, it is known that a second-order time differential equation can be transformed into a first-order differential equation in $(\dot{s}^2, s)$ based on the relation introduced in \cite{pfeiffer1987concept},
\begin{eqnarray}
\ddot{s} = \frac{d\dot{s}}{dt} =  \frac{d\dot{s}}{ds}\frac{ds}{dt} = \dot{s}'\dot{s}=\frac{1}{2}(\dot{s}^2)'.
\label{eqn:uandx}
\end{eqnarray}
This is beneficial for solving TOTP problems, as it allows writing all the equations in linear form by setting ($\dot{s}^2, \ddot{s}$) as optimization variables \cite{pham2018new, verscheure2009time}.

\subsection{TOTP Constraints in the Discretized System}

We now show that we can express all constraints up to second-order as linear in $x=\dot{s}^2$ for the discretized system. By dividing the path interval $[0,1]$ into $N$ segments, we have
\begin{eqnarray*}
0=:s_0, s_1,...,s_{N-1},s_N:=1.
\end{eqnarray*}
Then from the relation in \eqref{eqn:uandx}, $\dot{s}_k, \ddot{s}_k$ can be represented as:
\begin{equation}
\dot{s}_k^2 := x_k, \quad \ddot{s}_k = \frac{x_{k+1}-x_k}{2\triangle_k} \quad\textrm{or}\quad \frac{x_{k}-x_{k-1}}{2\triangle_{k-1}} \label{eqn:def_sdot_sddot},
\end{equation}
where $\triangle_k :=s_{k+1}-s_k$ and $\triangle_{k-1} :=s_{k}-s_{k-1}$.

For representational convenience, we simplify the notation of each quantity at $s_k$ as $(\cdot)(s_k)=(\cdot)_k$. 
Instead of introducing another optimization variable $u=\ddot{s}$ like other works~\cite{pham2018new}, we express all quantities as linear in $x$, which is sufficient. Finally, we formulate the following constraints:
\begin{itemize}
\item \textbf{Velocity limits (1st-order constraints)}
 First, the joint velocity limit can be represented as
\begin{eqnarray}
     \dot{\bq}_k \circ \dot{\bq}_k  = (\bq_k' \circ \bq_k') x_k &\leq& \dot{\bq}_{k}^{max} \circ \dot{\bq}_{k}^{max}, \label{eqn:1stvel}
\end{eqnarray}
where $\mathbf{a} \circ \mathbf{b}$ represents the Hadamard product (i.e., the element-wise product) of two vectors.
\item \textbf{Acceleration limits (2nd-order constraints)} Similarly, from the parameterized expression of acceleration 
$$
 \ddot{\bq}_k  = \bq''_k\dot{s_k}^2 + \bq'_k\ddot{s}_k =  \bq''_k x_k + \bq'_k\left(\frac{x_{k+1}-x_k}{2\triangle_k}\right),
$$
we can constrain two consecutive $x_k$ by
\begin{align}
  \left|  \frac{\bq'_k}{2\triangle_k} x_{k+1} + \left(\bq''_k - \frac{\bq'_k}{2\triangle_k}\right) x_k \right| \leq \ddot{\bq}_k^{max} & \nonumber \\ 
\textrm{or}\quad \left|  \left(\bq''_k + \frac{\bq'_k}{2\triangle_{k-1}}\right) x_k - \frac{\bq'_k}{2\triangle_{k-1}} x_{k-1} \right| \leq \ddot{\bq}_k^{max}. &
 \label{eqn:2ndacc}
\end{align}
\item \textbf{Torque limits (2nd-order constraints)}  
The parameterized dynamics can also be rewritten with respect to $x$ by substituting \eqref{eqn:def_sdot_sddot} into \eqref{eqn:parameterized_dyn}:
\begin{align}
 \left| \frac{\mathbf{m}_k}{2\triangle_k}x_{k+1}  + \left(\mathbf{c}_k-\frac{\mathbf{m}_k}{2\triangle_k}\right) x_k + \mathbf{g}_k \right| &\leq \btau_k^{max}  \label{eqn:2ndtrq}  \nonumber  \\
 \textrm{or}~ \left| \left(\mathbf{c}_k+\frac{\mathbf{m}_k}{2\triangle_{k-1}}\right)x_{k}  - \frac{\mathbf{m}_k}{2\triangle_{k-1}} x_{k-1} + \mathbf{g}_k \right| &\leq \btau_k^{max}. 
\end{align}
\end{itemize}

By stacking up all the constraints derived above, we can simply express first-order constraints as $\bomega_k x_k \leq \bnu_k$, where $\bomega_k =  \bq_k' \circ \bq_k' $ and $\bnu_k = \dot{\bq}_{k}^{max} \circ \dot{\bq}_{k}^{max}$. Similarly, second-order constraints can be represented in the form of $\balpha_k^0 x_k + \balpha_k^1 x_{k+1} \leq \bbeta_k$.

\section{Jerk-Constrained Time-Optimal Trajectory Planning}
\label{sec:jerktotp}
\subsection{Jerk Constraint Formulation}
Suppose we want to enforce joint-level jerk bounds
\begin{equation} \label{eqn:jerkbound}
-\dddot{\bq}_\textrm{max} \leq \dddot{\bq} \leq \dddot{\bq}_\textrm{max} .
\end{equation}
In this paper, we define jerk as change in acceleration between adjacent trajectory points. While this formula is more forgiving than an instantaneous rate, it is still effective in preventing abrupt changes in acceleration and simplifies the formulation. Additional details can be found in Section~\ref{sec:conclusion}. Then joint jerk given the discretized system can be formulated as: 
\begin{align}
\dddot{\bq}_k = \frac{\Delta \ddot{\mathbf{q}}}{\Delta t} (s_k)  &= \frac{\left(\ddot{\bq}_{k+1}-\ddot{\bq}_k\right)}{\frac{1}{2}(\triangle t_{k+1} + \triangle t_k)} \nonumber \\
 &= \frac{\mathbf{j}_k^2x_{k+2} + \mathbf{j}_k^1x_{k+1} + \mathbf{j}_k^0x_{k}}{h_k(\mathbf{x}_{k:k+2})} \label{eqn:jerk},
\shortintertext{where}
 \mathbf{x}_{k:k+2} &= [x_k ~ x_{k+1} ~ x_{k+2}]^\top \nonumber \\
h_k(\mathbf{x}_{k:k+2}) &=  {\dfrac{\triangle_{k+1}}{\sqrt{x_{k+2}}+\sqrt{x_{k+1}}}
 +\dfrac{\triangle_{k}}{\sqrt{x_{k+1}}+\sqrt{x_{k}}}} \nonumber 
 \end{align}
 \vspace{-4mm}
  {\footnotesize{
 \begin{align*}
\mathbf{j}_k^2 =\frac{\bq'_{k+1}}{2\triangle_{k+1}}&, \quad \mathbf{j}_k^1 = \bq''_{k+1}-\frac{\bq'_{k+1}}{2\triangle_{k+1}}-\frac{\bq'_{k}}{2\triangle_{k}},  
 \mathbf{j}_k^0 &=  - \bq_k''+\frac{\bq_k'}{2\triangle_k} \nonumber. 
\end{align*}
}}

Unfortunately, unlike second- or lower-order constraints, a nonlinear term such as $h_k(\mathbf{x}_{k:k+2})$ in Eqn. \eqref{eqn:jerk} always remains when expressing jerk in the TOTP problem. We address this issue by linearizing the nonlinear term $h_k(\mathbf{x}_{k:k+2})$ and leveraging its convexity to ensure the solution always satisfies the original constraints, i.e., 

\small
\begin{align*}
h_k(\mathbf{x}_{k:k+2}) &\geq h_k(\bar{\mathbf{x}}_{k:k+2}) + \nabla h_k(\bar{\mathbf{x}}_{k:k+2})({\mathbf{x}}_{k:k+2} - \bar{\mathbf{x}}_{k:k+2}) \\
&:= \bar{h}_k + \bar{h}_k^2x_{k+2} + \bar{h}_k^1x_{k+1} + \bar{h}_k^0x_{k},
\end{align*}
\normalsize
where
\begin{align*}
\bar{h}_k^i&=\frac{\partial h_k(\mathbf{x}_{k:k+2})}{\partial x_{k+i}}\bigg|_{\mathbf{x}=\mathbf{\bar{x}}}, \quad  i \in \{0,1,2\},\\
\bar{h}_k &=h_k(\bar{\mathbf{x}}_{k:k+2}) - \bar{h}_k^2 \bar{x}_{k+2}-\bar{h}_k^1 \bar{x}_{k+1}-\bar{h}_k^0 \bar{x}_{k}.
\end{align*}
We show the convexity of this function in the Appendix. From this, we can reformulate the joint jerk constraints as
\begin{align*}
\big|~ \mathbf{j}_k^2x_{k+2} &+ \mathbf{j}_k^1x_{k+1} + \mathbf{j}_k^0x_{k} ~\big| \\
&\leq  \dddot{\bq}_\textrm{max} \left( \bar{h}_k + \bar{h}_k^2x_{k+2} + \bar{h}_k^1x_{k+1} + \bar{h}_k^0x_{k} \right) \\
&\leq \dddot{\bq}_\textrm{max} h_k(\mathbf{x}_{k:k+2}) ,
\end{align*}
which finally leads to the linear form of 3rd-order constraints:
$$\bgamma_k^0 x_k + \bgamma_k^1 x_{k+1} + \bgamma_k^2 x_{k+2} \leq \boeta_k.$$
We can now combine all first- to third-order constraints as a linear matrix inequality:
\small
\begin{numcases}{}
\textrm{1st order constraints:}&{$\bomega_k x_k \leq \bnu_k$} \nonumber\\
\textrm{2nd order constraints:}&{$\balpha_k^0 x_k + \balpha_k^1 x_{k+1} \leq \bbeta_k$ }\label{eqn:1and2and3} \\
\textrm{3rd order constraints:}&$\bgamma_k^0 x_k + \bgamma_k^1 x_{k+1} + \bgamma_k^2 x_{k+2} \leq \boeta_k$ \nonumber
 \end{numcases}
\normalsize
\begin{figure*}[t!]
    \centering
    \vspace{2mm}
    \includegraphics[width=\linewidth]{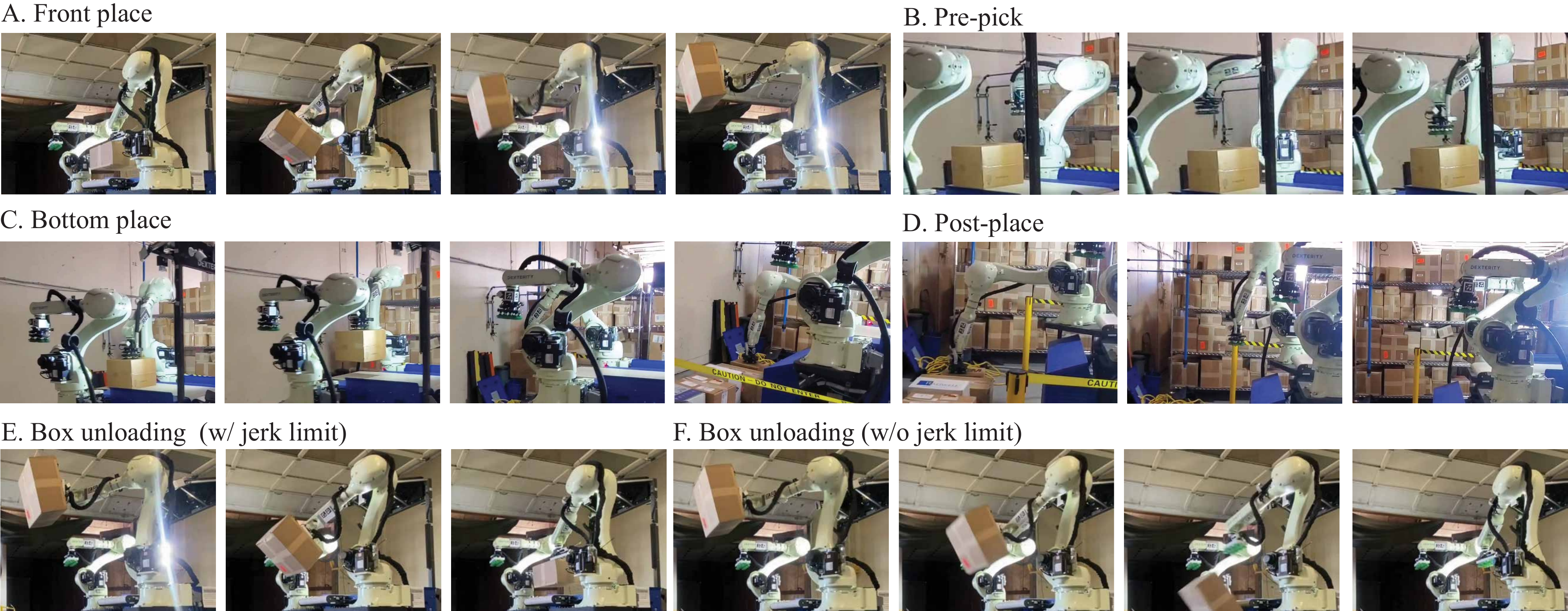}
    \caption{Test motion snapshots. A: Front place motion to move boxes from a conveyor to the top of a virtual stack of packages. B: Pre-pick motion preparing to pick a box from a conveyor: C. Bottom place motion to place boxes onto the bottom of the truck. D: Post-place motion to return to the ready pose after placing boxes. E and F show box unloading motions, which return the box from a stack of packages to the conveyor with and without jerk constraints. Without jerk constraints, the robot's suction gripper is also more likely to lose grasp of the box.}
    \label{fig:snapshot}
    \vspace{-10pt}
\end{figure*}

\subsection{Trajectory Optimization}
The cost function for minimizing the total time required to follow the given path can be expressed as
\begin{equation}
    f(\mathbf{x}) = \sum_{i=0}^{N-1} \frac{s_{i+1}-s_{i}}{\sqrt{x_i}+\sqrt{x_{i+1}}},
\end{equation}
which can be linearized along the nominal trajectory $\bar{\mathbf{x}}$. 

Now we formulate the trajectory optimization problem as a Sequential Linear Program (SLP), where the optimization variables are $\mathbf{x} = [x_1, \cdots, x_{N-1}]^\top$ as follows: 
\begin{align}
& \min_{\mathbf{x}} \quad \mathbf{c}^\top\mathbf{x}  \label{eqn:opt} \\
& \textrm{subject to}\quad A \mathbf{x} \leq b, \nonumber
\end{align}
where
\begin{equation}
\mathbf{c} = \frac{\partial f}{\partial \mathbf{x}}\bigg\vert_{\mathbf{x}=\mathbf{\bar{x}}} \label{eqn:constraints}, \; A = \begin{bmatrix}  A^1 \\ A^2 \\ A^3  \end{bmatrix},\; b = \begin{bmatrix}  b^1 \\ b^2 \\ b^3 \end{bmatrix} \nonumber
\end{equation}  \vspace{-10pt}
{\footnotesize
\begin{align}
A^1 &= \begin{bmatrix}
\bomega_1 & 0 & \cdots & 0 \\
\vdots & &  & \vdots \\
 0 & \cdots & 0 & \bomega_{N-1} 
\end{bmatrix}, %
\quad\;\; b^1 = \begin{bmatrix}
\bnu_1 \\ \vdots \\ \bnu_{N-1} 
\end{bmatrix}, \nonumber \\
A^2 &= \begin{bmatrix}
\balpha^1_{0} & 0 & \cdots & 0 \\
\balpha^0_{1} & \balpha^1_{1} & \cdots & 0 \\
\vdots & & & \vdots \\
0 & \cdots & \balpha^0_{N-2} & \balpha^1_{N-2}   \\
0 & \cdots & \cdots & \balpha^0_{N-1}   
\end{bmatrix}, %
b^2 = \begin{bmatrix}
\bbeta_{0}-\balpha_0^0x_0 \\
\bbeta_{1} \\ \vdots \\ \bbeta_{N-2} \\ 
\bbeta_{N-1} - \balpha_{N-1}^1x_N
\end{bmatrix}, \nonumber \\
A^3 &= \begin{bmatrix}
\bgamma^2_0 & 0 & 0 & 0 & \cdots & 0 \\
\bgamma^1_1 & \bgamma^2_1 & 0 & 0 & \cdots & 0 \\
\bgamma^0_2 & \bgamma^1_2 & \bgamma^2_2 & 0 & \cdots & 0 \\
\vdots & & \vdots & \ddots & \ddots & \vdots \\
0 & \cdots & 0 & \bgamma^0_{N-3} & \bgamma^1_{N-3} & \bgamma^2_{N-3}\\
0 & \cdots & 0 & 0 & \bgamma^0_{N-2} & \bgamma^1_{N-2}
\end{bmatrix}, \label{eqn:A3} \\
b^3 &= \begin{bmatrix}  \boeta_0 - \bgamma^0_0 x_0 -\bgamma^1_0 x_1 \\
 \boeta_{1} - \bgamma^0_{1}x_1 \\
 \boeta_{2} \\ \vdots \\ \boeta_{N-3} \\ \boeta_{N-2} - \bgamma^2_{N-2} x_N
\end{bmatrix}.
\label{eqn:b3}
\end{align}
}
\vspace{3mm}

Note that due to the linearization, we need to iteratively update the nominal trajectory to reformulate the coefficients for cost function and 3rd-order constraints. The steps of the resulting algorithm are shown in Algorithm~\ref{alg:cap}.

\begin{algorithm}
\caption{TOTP with 3rd-order constraints (TOTP3)}\label{alg:cap}
\begin{algorithmic}
\Require Nominal variables $\bar{\mathbf{x}}_{1:N}$
\While {True}
\State Update $\mathbf{c}, A^3, b^3$ \Comment{Eqn. \eqref{eqn:constraints},\eqref{eqn:b3}}
\State $\mathbf{x}_{1:N}$ = solve LP \Comment{Eqn. \eqref{eqn:opt}}
\If {$\|\mathbf{x}_{1:N}-\bar{\mathbf{x}}_{1:N}\| < \epsilon$} break
\EndIf
\State Update $\bar{\mathbf{x}}_{1:N}$
\EndWhile
\end{algorithmic}
\end{algorithm}

\begin{figure}[b!]
    \centering    
    \vspace{-10pt}
    \includegraphics[width=0.85\linewidth]{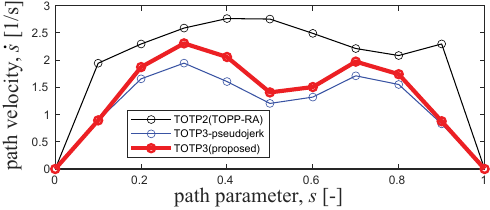}
    \vspace{-10pt}
    \caption{Comparison of the velocity curve for the pre-pick motion (Fig.~\ref{fig:snapshot}B) computed from TOPP-RA \cite{pham2018new} without jerk limits (black), \cite{zhang2013practical} with pseudo jerk limits (blue) and the proposed approach with jerk limits (red).}
    \label{fig:s_sdot}
\end{figure}


\section{EXPERIMENT RESULTS}
\label{sec:experiment}
In this section, we provide experimental results to validate the effectiveness of the proposed approach.

\begin{figure*}[ht!]
    \centering
    \vspace{2mm}
    \includegraphics[width=\linewidth]{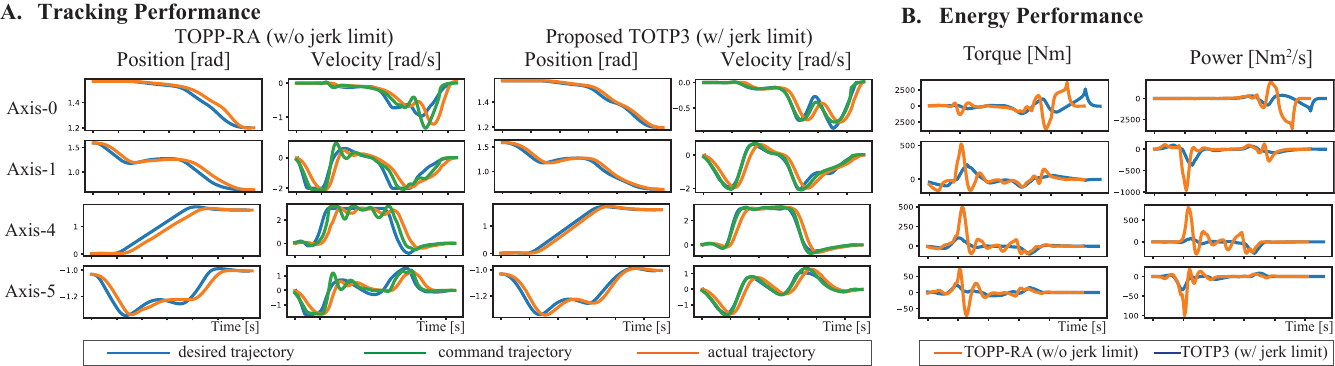}
    \vspace{-20pt}
    \caption{Algorithm performance comparison on real-robot for the front place motion (Fig.~\ref{fig:snapshot}A): \textbf{A} shows tracking performance of the algorithms with and without jerk limits by comparing the desired, commanded, and actual joint position and velocity. \textbf{B} shows the corresponding estimated torque and power. }
    \label{fig:performance}
    \vspace{-5pt}
\end{figure*}

\begin{figure}[ht!]
    \centering    
    \includegraphics[width=\linewidth]{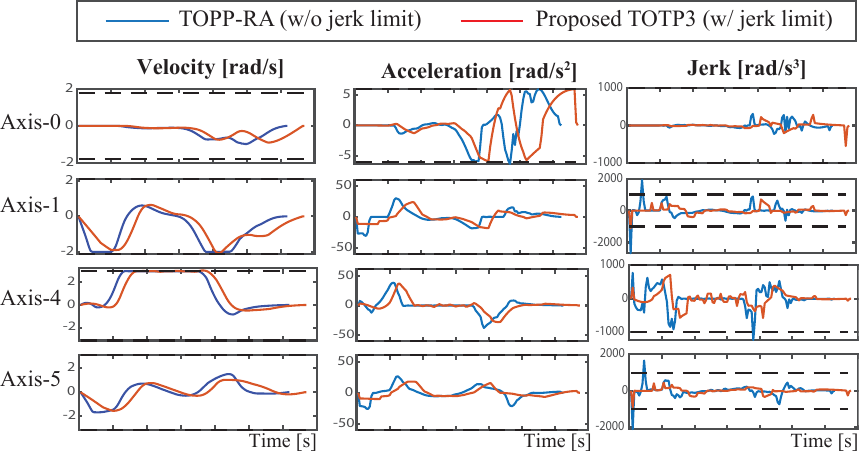}
    \vspace{-20pt}
    \caption{Time-optimal trajectory generated for the front place motion (Fig.~\ref{fig:snapshot}A) without jerk limits (blue) and with jerk limits (red).}
    \label{fig:jerkplot}
    \vspace{-10pt}
\end{figure}

\subsection{Experiment Setup}
 For practical evaluation, we used an example motion from a real-world box-loading application. We performed the test motions depicted and described in Fig.~\ref{fig:snapshot} with and without the jerk limit. We used the output from TOPP-RA \cite{pham2018new} as the initial nominal trajectory for our algorithm and as a comparison target for the optimal trajectory without the jerk limit. Finally, the jerk limits were heuristically chosen to be $100 \sim 1000 
 \textnormal{ rad}/\textnormal{s}^3$, balancing the trade-off between motion duration and overall performance metrics such as energy efficiency and tracking (note that some applications may require more specific strict limitations on jerk). Lastly, we used 7-DOF robotic system composed of a 6-DOF RS020N Kawasaki arm mounted on an additional revolute joint for our experiment. 

\subsection{Comparison With TOPP-RA }
\subsubsection{Simulation Results}
We compare the TOTP outcomes obtained by three different methods in the phase plane. These methods include TOPP-RA \cite{pham2018new} without jerk limits, TOTP3 \cite{zhang2013practical} with pseudo jerk limits, and our proposed approach with jerk limits. While the original work in \cite{zhang2013practical} focused on torque rate constraints, we adapted it for our experiment by introducing jerk limits. We employed a similar methodology to the one used for pseudo-torque rate to establish the pseudo-jerk limit.  Fig.~\ref{fig:s_sdot} shows that the curve tends to have a smoother shape when jerk limits are enforced. Note that since we formulate jerk limits in joint space and then map them to the space of the path parameter $s$, the smoothness of the joint-space curves along the path may vary at different path points. The result shows that the proposed approach yields faster motion than the pseudo-torque rate method under the same limits. To highlight the optimality gap, we applied a very stringent jerk limit of $100 \textnormal{ rad}/\textnormal{s}^3$. 

Fig.~\ref{fig:jerkplot} illustrates the computed velocity, acceleration, and jerk with the corresponding limits for the significant axes of the robot. It demonstrates that by enforcing a jerk limit, rapid changes in acceleration are mitigated, leading to a smoother velocity profile.

\subsubsection{Robot Implementation Results}

While we can observe jerk limits producing a smoother trajectory in simulation, we must still establish their tangible benefits on a real robot. We performed a comparative test on the real Kawasaki robot system described above, running the same path with and without jerk limits. The results are shown in Fig.~\ref{fig:performance}. 

Fig.~\ref{fig:performance}A shows that the robot deviated from its desired path during rapid acceleration changes, leading to overshooting and increased fluctuations in the velocity plots. However, applying jerk limits significantly smoothed the motions, resulting in differences in the generated torque and power, as illustrated in Fig.~\ref{fig:performance}B. The jerk-limited motions prevented aggressive velocity spikes, lowering peak power and torque usage, with significant reductions for some joints.

We measured peak power and RMS torque and the results are presented in Table~\ref{tab:energy}, showing that peak power was reduced by about 25\%, and RMS torque was reduced to half of its original value by limiting jerk.

\renewcommand{\arraystretch}{1.2}
\setlength{\tabcolsep}{5pt}

\begin{table}[t!]
\caption{RMS torque and peak power during front place motion}
\vspace{-2mm}
\centering
\begin{tabular}
{|c|c|c|c|c|c|c|c|c|}
\cline{3-9}
\multicolumn{2}{c|}{} & \multicolumn{7}{c|}{Axis} \\ \cline{3-9}
 \multicolumn{2}{c|}{} & \multicolumn{1}{c|}{0} & \multicolumn{1}{c|}{1} & \multicolumn{1}{c|}{2} & \multicolumn{1}{c|}{3} & \multicolumn{1}{c|}{4} & \multicolumn{1}{c|}{5} & \multicolumn{1}{c|}{6} \\ \hline
\multirow{2}{*}{\begin{tabular}[c]{@{}c@{}} {RMS Trq}\\ ~[Nm] \end{tabular}} & {TOPP-RA} & 3624 & 965 & 32 & 73 & 795 & 100 & 28 \\ \hhline{|~|--------}
 & TOTP3 & \cellcolor{LightCyan}1504 & \cellcolor{LightCyan}377 & 14 & 77 & \cellcolor{LightCyan}267 & 35 & 6.4 \\ \hline
\multirow{2}{*}{\begin{tabular}[c]{@{}c@{}} {Peak Pwr}\\ ~[$\textnormal{Nm}^2$/s] \end{tabular}} & {TOPP-RA} & 1119 & 113 & 57 & 40 & 102 & 18 & 8.5 \\ \hhline{|~|--------}
  & TOTP3 & \cellcolor{LightCyan}612 &\cellcolor{LightCyan}67 & 27 & 38 & \cellcolor{LightCyan}34 & 6.6 & 1.2 \\ \hline
\end{tabular}
\label{tab:energy}
\vspace{-10pt}
\end{table}



\subsubsection{Algorithm Efficiency}


We ran experiments via a C++ implementation running on a 64-bit system with a 2.35 GHz AMD EPYC 7452 32-core processor. To show the algorithm's efficiency, we measured the computation time of TOTP3 and TOPP-RA by averaging 20 trials of box-place motions. Each trial involved 20-30 path points beginning from slightly different initial box lift poses. TOTP3 (w/ jerk limit) took $7.533 \pm 5.848$~ms while TOPP-RA (w/o jerk limit) took $0.273 \pm 0.089$~ms to obtain the solution. The TOPT3 algorithm's runtime was longer than TOPP-RA due to the additional complexity of enforcing third-order constraints. However, it could still be executed within 10~ms, very reasonable for real-time applications. We were able to accelerate our computation times by warm-starting our algorithm with the non-jerk-limited TOPP-RA result and recomputing the only changed portion at each iteration. 

\begin{figure}[t!]
    \centering    
    \vspace{2mm}
    \includegraphics[width=0.9\linewidth]{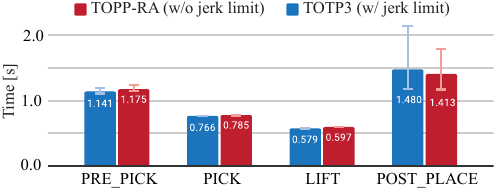}
    \vspace{-5pt}
    \caption{Motion duration for several motions computed with and without jerk limit.}
    \label{fig:motion_duration}
    \vspace{-10pt}
\end{figure}

Lastly, we measured the motion time generated by each algorithm for several different motions. Notably, imposing a jerk limit of $1000 \textnormal{ rad}/\textnormal{s}^3$  increased overall motion duration by just 3-5\%. For the ``Post-Place" motion, we observed lower motion time with jerk limit, but this is just due to a known case of suboptimality in TOPP-RA, which guarantees optimality only if path points exclude zero-inertia point (see \cite{pham2018new} for details), which requires denser path segmentation. On the other hand, the proposed algorithm's SLP formulation allows it to converge to the optimal solution.

\section{DISCUSSION \& CONCLUSIONS}
\label{sec:conclusion}
In summary, we introduced a novel approach to solve the time-optimal trajectory planning problem while enforcing third-order constraints. In particular, we constrained the acceleration change between path points to produce smoother and more trackable trajectories. Using the proposed formulation, we removed the need for high-order dynamics variables which are computationally difficult to handle. For instance, the jerk formulation in~\cite{pham2018new} requires $\mathbf{q}'''$, and the torque rate formulation in~\cite{zhang2013practical} requires $\mathbf{M}'$ and $\mathbf{C}'$. These quantities can be challenging to compute from standard robot dynamics libraries. Though this paper tested only jerk limits, the provided formulation is versatile and can be extended to other useful third-order constraints such as torque rate.

Furthermore, our study has shown a significant reduction in peak power and torque efficiency from these jerk limits. This comes at the expense of some increased computation time, though it remains well within real-time limits. Additionally, this trade-off can be managed by computing the next step while the robot is executing the preceding path and using warm-starting to recompute solutions as needed. Additionally, although jerk limits result in slightly longer motion durations, this can be compensated by reduced time required for the robot to stabilize and come to a stop, due to reduced tracking fluctuation and overshoot.  

During this study, we also observed that the results were quite sensitive to path parameter function discretization and splining.  Splines can interpolate the discretized solution to generate robot commands at the desired high control frequency, but some spline types may induce jerk discontinuities between spline segments. This can be mitigated by constraining acceleration changes in transitions between spline segments, which can be formulated as additional third-order constraints. Furthermore, it is crucial to ensure sufficient discretization resolution to avoid high jerk between knot points where jerk is not directly constrained. In the future, we should explore more optimal and constrained parameterization methods for TOTP's path parameter function and refine the overall optimization process to accommodate corresponding changes effectively.


\section*{APPENDIX}
\subsection{Proof of convexity}

\textbf{Proposition 1.} $f(x) = \dfrac{a}{\sqrt{x_1}+\sqrt{x_2}}+\dfrac{b}{\sqrt{x_2}+\sqrt{x_3}}$ \textit{is convex in $x=(x_1,x_2,x_3)\in \mathbb{R}_+^3$.}

\textit{Proof.} To show a twice-differentiable function $f$ is convex on a convex set, it is sufficient to show that its Hessian matrix of second partial derivatives is positive semi-definite in the interior of the set \cite{boyd2004convex}. The Hessian matrix is
\begin{align*}
    H &= \frac{\partial^2 f}{\partial x^2} = 
    \begin{bmatrix} s_{11} & s_{12} & 0 \\
        s_{12} & s_{22} & s_{23} \\
        0 & s_{23} & s_{33}
    \end{bmatrix} ,
\end{align*}
where
\footnotesize{
\begin{align*}
    s_{11} =& \frac{a}{2x_1(\sqrt{x_1}+\sqrt{x_2})^2}\left(\frac{1}{2\sqrt{x_1}}+\frac{1}{\sqrt{x_1}+\sqrt{x_2}}\right) \\    
    s_{22} =& \frac{a}{2x_2(\sqrt{x_1}+\sqrt{x_2})^2}\left(\frac{1}{\sqrt{x_1}+\sqrt{x_2}}+\frac{1}{2\sqrt{x_2}}\right) \\
    &+ \frac{b}{2x_2(\sqrt{x_2}+\sqrt{x_3})^2}\left(\frac{1}{2\sqrt{x_2}}+\frac{1}{\sqrt{x_2}+\sqrt{x_3}}\right) \\
    s_{33} =& \frac{b}{2x_3(\sqrt{x_2}+\sqrt{x_3})^2}\left(\frac{1}{\sqrt{x_2}+\sqrt{x_3}}+\frac{1}{2\sqrt{x_3}}\right) \\
    s_{12} =& \frac{a}{2\sqrt{x_1}\sqrt{x_2}(\sqrt{x_1}+\sqrt{x_2})^3} , \;
    s_{23} = \frac{b}{2\sqrt{x_2}\sqrt{x_3}(\sqrt{x_2}+\sqrt{x_3})^3}. 
\end{align*}
}

\normalsize{
As $H$ is a real symmetric matrix, $\lambda$s (eigenvalues of $H$) are real, which can be obtained from the characteristic equation: }
\begin{align*}
    \lambda^3 - \text{Tr}(H)\lambda^2 + \Bigg(\sum_{i \neq  j}s_{ii}s_{jj} - s_{12}^2 -s_{23}^2\Bigg)\lambda - |H| = 0.
\end{align*} 
First, given the domain of $x_i>0$, it is clear that $\text{Tr}(H)>0$ and $\sum_{i \neq j}s_{ii}s_{jj} - s_{12}^2 -s_{23}^2 >0 $. The determinant of $H$ is also positive in the given domain:

\scriptsize{
\begin{align*}
    |H| = \dfrac{3ab(a(4\sqrt{x_2}\sqrt{x_3}+x_2+3x_3)+b(4\sqrt{x_1}\sqrt{x_2}+3x_1+x_2))}{64\sqrt{x_1^3 x_2^3 x_3^3}(\sqrt{x_1}+\sqrt{x_2})^4(\sqrt{x_2}+\sqrt{x_3})^4  } > 0. 
\end{align*} 
}
\normalsize{
When considering the relationship between the roots and coefficients of a cubic equation, it is known that if all coefficients above are positive, then all $\lambda$s satisfying the equation are also positive. Thus we have shown that $H$ is positive semi-definite, which implies that $f$ is convex.
}

\vspace{5mm}

\section*{ACKNOWLEDGMENTS}
We thank Dexterity and the HCRL personnel for their support of this project. Jee-eun Lee was a robotics intern for Dexterity, Inc. during the summer of 2023, and Luis Sentis was a consultant for Dexterity, Inc. during the summer of 2022.


\vspace{5mm}





\bibliographystyle{bib/IEEEtran}
\bibliography{bib/root}

\begin{thebibliography}{10}
\providecommand{\url}[1]{#1}
\csname url@rmstyle\endcsname
\providecommand{\newblock}{\relax}
\providecommand{\bibinfo}[2]{#2}
\providecommand\BIBentrySTDinterwordspacing{\spaceskip=0pt\relax}
\providecommand\BIBentryALTinterwordstretchfactor{4}
\providecommand\BIBentryALTinterwordspacing{\spaceskip=\fontdimen2\font plus
\BIBentryALTinterwordstretchfactor\fontdimen3\font minus \fontdimen4\font\relax}
\providecommand\BIBforeignlanguage[2]{{%
\expandafter\ifx\csname l@#1\endcsname\relax
\typeout{** WARNING: IEEEtran.bst: No hyphenation pattern has been}%
\typeout{** loaded for the language `#1'. Using the pattern for}%
\typeout{** the default language instead.}%
\else
\language=\csname l@#1\endcsname
\fi
#2}}

\bibitem{chettibi2004minimum}
T.~Chettibi, H.~Lehtihet, M.~Haddad, and S.~Hanchi, ``Minimum cost trajectory planning for industrial robots,'' \emph{European Journal of Mechanics-A/Solids}, vol.~23, no.~4, pp. 703--715, 2004.

\bibitem{diehl2006fast}
M.~Diehl, H.~G. Bock, H.~Diedam, and P.-B. Wieber, ``Fast direct multiple shooting algorithms for optimal robot control,'' \emph{Fast Motions in Biomechanics and Robotics}, pp. 65--93, 2006.

\bibitem{schulman2013finding}
J.~Schulman, J.~Ho, A.~X. Lee, I.~Awwal, H.~Bradlow, and P.~Abbeel, ``Finding locally optimal, collision-free trajectories with sequential convex optimization.'' in \emph{Robotics: Science and Systems}, vol.~9, no.~1.\hskip 1em plus 0.5em minus 0.4em\relax Berlin, Germany, 2013.

\bibitem{zhao2018efficient}
Y.~Zhao, H.-C. Lin, and M.~Tomizuka, ``Efficient trajectory optimization for robot motion planning,'' in \emph{International Conference on Control, Automation, Robotics and Vision (ICARCV)}.\hskip 1em plus 0.5em minus 0.4em\relax IEEE, 2018, pp. 260--265.

\bibitem{zhang2022time}
X.~Zhang, F.~Xiao, X.~Tong, J.~Yun, Y.~Liu, Y.~Sun, B.~Tao, J.~Kong, M.~Xu, and B.~Chen, ``Time optimal trajectory planning based on improved sparrow search algorithm,'' \emph{Frontiers in Bioengineering and Biotechnology}, vol.~10, 2022.

\bibitem{shin1985minimum}
K.~Shin and N.~McKay, ``Minimum-time control of robotic manipulators with geometric path constraints,'' \emph{IEEE Transactions on Automatic Control}, vol.~30, no.~6, pp. 531--541, 1985.

\bibitem{shiller1989robot}
Z.~Shiller and S.~Dubowsky, ``Robot path planning with obstacles, actuator, gripper, and payload constraints,'' \emph{The International Journal of Robotics Research}, vol.~8, no.~6, pp. 3--18, 1989.

\bibitem{shiller1991computing}
------, ``On computing the global time-optimal motions of robotic manipulators in the presence of obstacles,'' \emph{IEEE Transactions on Robotics and Automation}, vol.~7, no.~6, pp. 785--797, 1991.

\bibitem{bobrow1985time}
J.~E. Bobrow, S.~Dubowsky, and J.~S. Gibson, ``Time-optimal control of robotic manipulators along specified paths,'' \emph{The international journal of robotics research}, vol.~4, no.~3, pp. 3--17, 1985.

\bibitem{pfeiffer1987concept}
F.~Pfeiffer and R.~Johanni, ``A concept for manipulator trajectory planning,'' \emph{IEEE Journal on Robotics and Automation}, vol.~3, no.~2, pp. 115--123, 1987.

\bibitem{eager2016beyond}
D.~Eager, A.-M. Pendrill, and N.~Reistad, ``Beyond velocity and acceleration: {Jerk,} snap and higher derivatives,'' \emph{European Journal of Physics}, vol.~37, no.~6, 2016.

\bibitem{pham2014general}
Q.-C. Pham, ``A general, fast, and robust implementation of the time-optimal path parameterization algorithm,'' \emph{IEEE Transactions on Robotics}, vol.~30, no.~6, pp. 1533--1540, 2014.

\bibitem{verscheure2009time}
D.~Verscheure, B.~Demeulenaere, J.~Swevers, J.~De~Schutter, and M.~Diehl, ``Time-optimal path tracking for robots: A convex optimization approach,'' \emph{IEEE Transactions on Automatic Control}, vol.~54, no.~10, pp. 2318--2327, 2009.

\bibitem{hauser2014fast}
K.~Hauser, ``Fast interpolation and time-optimization with contact,'' \emph{The International Journal of Robotics Research}, vol.~33, no.~9, pp. 1231--1250, 2014.

\bibitem{shin1986dynamic}
K.~Shin and N.~McKay, ``A dynamic programming approach to trajectory planning of robotic manipulators,'' \emph{IEEE Transactions on Automatic Control}, vol.~31, no.~6, pp. 491--500, 1986.

\bibitem{singh1987optimality}
C.~Singh, ``Optimality conditions in multiobjective differentiable programming,'' \emph{Journal of Optimization Theory and Applications}, vol.~53, pp. 115--123, 1987.

\bibitem{haschke2008line}
R.~Haschke, E.~Weitnauer, and H.~Ritter, ``On-line planning of time-optimal, jerk-limited trajectories,'' in \emph{2008 IEEE/RSJ International Conference on Intelligent Robots and Systems}.\hskip 1em plus 0.5em minus 0.4em\relax IEEE, 2008, pp. 3248--3253.

\bibitem{berscheid2021jerk}
L.~Berscheid and T.~Kr{\"o}ger, ``Jerk-limited real-time trajectory generation with arbitrary target states,'' \emph{arXiv preprint arXiv:2105.04830}, 2021.

\bibitem{pham2018new}
H.~Pham and Q.-C. Pham, ``A new approach to time-optimal path parameterization based on reachability analysis,'' \emph{IEEE Transactions on Robotics}, vol.~34, no.~3, pp. 645--659, 2018.

\bibitem{consolini2019optimal}
L.~Consolini, M.~Locatelli, A.~Minari, A.~Nagy, and I.~Vajk, ``Optimal time-complexity speed planning for robot manipulators,'' \emph{IEEE Transactions on Robotics}, vol.~35, no.~3, pp. 790--797, 2019.

\bibitem{pham2017structure}
H.~Pham and Q.-C. Pham, ``On the structure of the time-optimal path parameterization problem with third-order constraints,'' in \emph{IEEE International Conference on Robotics and Automation (ICRA)}.\hskip 1em plus 0.5em minus 0.4em\relax IEEE, 2017, pp. 679--686.

\bibitem{ma2021new}
J.~Ma, S.~Gao, H.~Yan, Q.~Lv, and G.~Hu, ``A new approach to time-optimal trajectory planning with torque and jerk limits for robot,'' \emph{Robotics and Autonomous Systems}, vol. 140, 2021.

\bibitem{mattmuller2009calculating}
J.~Mattm{\"u}ller and D.~Gisler, ``Calculating a near time-optimal jerk-constrained trajectory along a specified smooth path,'' \emph{The International Journal of Advanced Manufacturing Technology}, vol.~45, pp. 1007--1016, 2009.

\bibitem{wang2021third}
M.~Wang, J.~Xiao, S.~Liu, and H.~Liu, ``A third-order constrained approximate time-optimal feedrate planning algorithm,'' \emph{IEEE Transactions on Robotics}, vol.~38, no.~4, pp. 2295--2307, 2021.

\bibitem{lu2017solving}
S.~Lu, J.~Zhao, L.~Jiang, H.~Liu, \emph{et~al.}, ``Solving the time-jerk optimal trajectory planning problem of a robot using augmented lagrange constrained particle swarm optimization,'' \emph{Mathematical Problems in Engineering}, vol. 2017, 2017.

\bibitem{zhang2013practical}
Q.~Zhang, S.~Li, and X.~Gao, ``Practical smooth minimum time trajectory planning for path following robotic manipulators,'' in \emph{American Control Conference}.\hskip 1em plus 0.5em minus 0.4em\relax IEEE, 2013, pp. 2778--2783.

\bibitem{boyd2004convex}
S.~P. Boyd and L.~Vandenberghe, \emph{Convex optimization}.\hskip 1em plus 0.5em minus 0.4em\relax Cambridge University Press, 2004.

\end{thebibliography}

\end{document}